\documentclass[journal]{IEEEtran}
\usepackage{booktabs}
\usepackage{multirow}
\usepackage{multirow}
\usepackage{longtable}
\usepackage{graphicx}
\usepackage{bm}
\usepackage{algorithm}
\usepackage[]{algpseudocode}
\usepackage[switch]{lineno}
\usepackage{caption}
\usepackage{color}

\usepackage{ifpdf}

\usepackage{cite}

\ifCLASSINFOpdf
\else
\fi
\usepackage{amsmath}
\usepackage{array}
\usepackage{fixltx2e}

\usepackage{stfloats}
\usepackage{url}

\begin{document}
%
\title{Predicting Events in MOBA Games: Prediction, Attribution, and Evaluation}
%
%
%

\author{Zelong~Yang,
        Yan~Wang$^\dag$,
        Piji~Li,
        Shaobin~Lin,
        Shuming~Shi,
        Shao-Lun~Huang,
        and~Wei~Bi
        
\thanks{$^\dag$:~corresponding author.}
\thanks{Zelong Yang (yangzelong14@gmail.com) and Shao-Lun Huang are with Tsinghua-Berkeley Shenzhen Institute, Tsinghua University, Shenzhen,
China. Yan Wang (brandenwang@tencent.com), Piji Li, Shaobin Lin, Shuming Shi, and Wei Bi are with Tencent AI Lab, Shenzhen, China.}  
\thanks{Manuscript received December 23, 2020; revised July 28, 2021, December 9, 2021 and February 20, 2022.}}

%
%

\markboth{Journal of \LaTeX\ Class Files,~Vol.~14, No.~8, August~2015}%
{Shell \MakeLowercase{\textit{et al.}}: Bare Demo of IEEEtran.cls for IEEE Journals}
%



\maketitle

\begin{abstract}
The multiplayer online battle arena (MOBA) games have become increasingly popular in recent years. Consequently, many efforts have been devoted to providing pre-game or in-game predictions for them.
\textcolor{black}{These predictions can be used in many MOBA esports related applications, such as AI commentator systems, in-game data analysis, and game-assistant bots.} However, these works are limited in the following two aspects: 1) the lack of sufficient in-game features; 2) the absence of interpretability in the prediction results. These two limitations greatly restrict the practical performance and industrial application of the current works. In this work, we collect a large-scale dataset containing rich in-game features for the popular MOBA game \emph{Honor of Kings}. \textcolor{black}{We then propose to predict four types of 
prediction tasks in an interpretable way by attributing the predictions to the input features using two gradient-based attribution methods: \textit{Integrated Gradients} and \textit{SmoothGrad}.} To evaluate the explanatory power of different models and attribution methods, a fidelity-based evaluation metric is further proposed. Finally, we evaluate the accuracy and Fidelity of several competitive methods to assess how well machines predict events in MOBA games. 

\end{abstract}


%
\IEEEpeerreviewmaketitle

\section{Introduction}

Nowadays, with the fast development of the gaming industry, electronic games are becoming increasingly popular, generating huge amounts of profit. Among all the genres of electronic games, MOBA games are one of the most popular and highest-grossing types, such as \emph{Defense of the Ancient II} \footnote{\url{https://en.wikipedia.org/wiki/Dota_2}\label{dota2}} (\emph{DotA2}), \emph{League of Legends}\footnote{\url{https://en.wikipedia.org/wiki/League_of_Legends}\label{lol}} (\emph{LoL}), and \emph{Honor of Kings}\footnote{\url{https://en.wikipedia.org/wiki/Honor_of_Kings}\label{hok}} (\emph{HoK}). Together, these three popular MOBA games have more than 300 million monthly active players\textsuperscript{\ref{dota2}\ref{lol}\ref{hok}} globally and an even larger potential audience among the streaming media community. Along with MOBA games' flourishing, much research has been done to predict the results before and during the games. These studies can be categorized into two types: pre-game predictions that predict the results of MOBA games before they begin \cite{conley2013does}, and in-game predictions that predict according to the in-game situations of the games \cite{yang2016real}. \textcolor{black}{These predictions can provide more information to the audience and the commentators, and therefore can be used in many MOBA related applications, such as the AI commentator systems, the in-game data analysis, and the game-assistant bots that can provide basic analysis and suggestions for the players.}

In many senses, the in-game predictions are more significant and have broader application scenarios than the pre-game predictions. However, although some progress has been made, existing studies for in-game predictions are limited in the following two aspects. The first limitation is: the insufficiency of large-scale in-game data. Due to the difficulty in collecting in-game data, current datasets for MOBA game predictions contain only pre-game features \cite{semenov2016performance} or limited types of in-game features such as ``gold'', ``experience'', and ``death'' \cite{yang2016real}. 

\textcolor{black}{Secondly, the predictions of the current works are non-interpretable, which greatly limits their application. The term ``non-interpretable'' means that the predictions are given without showing the underlying reasoning why these results are achieved by the prediction models. By showing what game features in the current input are contributing to the prediction result, the researchers and the audience can get more information that is helpful to understand the games.}
\begin{figure}[t]
\centering
\includegraphics[scale=0.30]{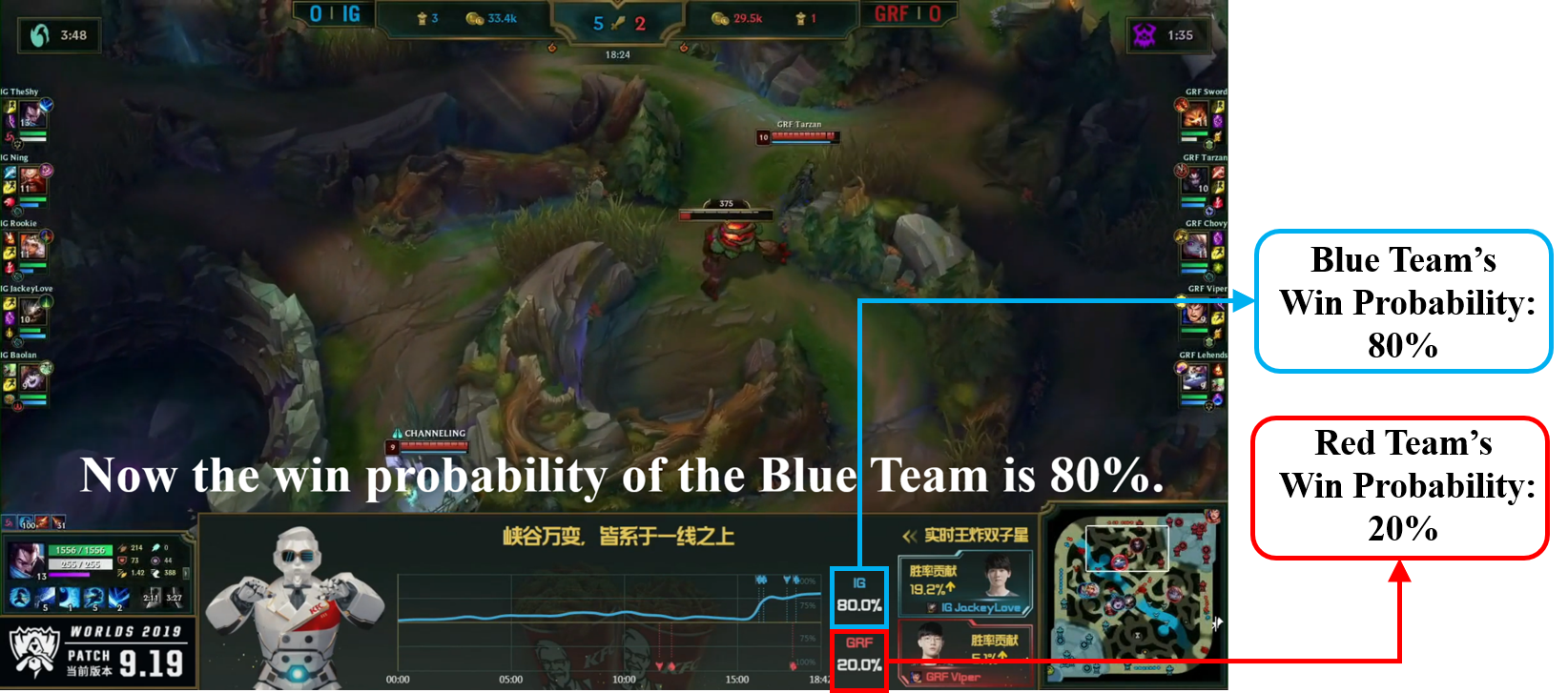}
\caption{Non-interpretable prediction in the LoL 2019 World Championship: the audience can only be informed of the winning probabilities of the two teams.}
\label{fig:lol4}
\end{figure}
As one can see in Figure \ref{fig:lol4}, the non-interpretable model used in \emph{LoL} 2019 World Championship can only give the winning probabilities of the two teams (80\% versus 20\%), which is opaque to the audience.\footnote{\textcolor{black}{Besides technical reasons, there may be other non-technical considerations that cause the non-transparency in official MOBA games, such as protecting proprietary knowledge.}} \textcolor{black}{The audience can understand the game situations better if they are shown what factors result in these predictions.} \textcolor{black}{For example, if Team-1 gets a larger winning rate and the feature ``campMoneyDiff'' (the difference between the two team's gold-amount) is one of the key features to the win prediction, then Team-1's superiority is (partly) due to its lead in gold-amount. Moreover, if a significant change appeared in the predictions within a short period (such as five seconds), then the top-contributing features will explain why this sudden breakthrough happens. From the perspective of both the researchers and the audience, this additional explanatory information is valuable and helpful to understand what is actually going on in the match, as human's analyses may not always be consistent with the prediction models' attributions, and when this happens, the interpretable results can give more information to the researchers and the audience.} Hence, there is a need to generate human-interpretable predictions for MOBA games.

In this work, to facilitate the study of in-game predictions for MOBA games, we collect a large-scale dataset that contains in-game records with rich features extracted from the game-core data (the back-end data of \emph{HoK}) of 50,278 games. \textcolor{black}{
Every second, we record more than 2,000 features. Within this dataset, one can easily train models to predict the important events of \emph{HoK}. 
}

Given the game records as input, we further train two state-of-the-art (SOTA) sequence modeling networks, Long Short-Term Memory network (LSTM) \cite{hochreiter1997long} and Transformer \cite{vaswani2017attention}, to predict the important events in \emph{HoK}. However, as noted above, these well-known black-box networks are non-interpretable, so one cannot understand the reasons for their predictions. To mitigate this problem, we propose to apply two gradient-based attribution methods, \textit{Integrated Gradients} (IG)~\cite{sundararajan2017axiomatic} and \textit{SmoothGrad} (SG)~\cite{smilkov2017smoothgrad}, to interpret the prediction result by attributing it to the top-contributed feature-dimensions. As reasoned by \cite{sundararajan2017axiomatic} and \cite{smilkov2017smoothgrad}, with these methods we can determine which features primarily contribute to the current prediction result.

Although the attribution methods can interpret the predictions of deep neural networks, there is another issue in that it is difficult to evaluate the correctness of the attribution methods. In other words, we can hardly know how well they interpret the prediction results. So, inspired by the fidelity-based evaluation methods in Natural Language Processing \cite{li2020evaluating, jacovi2020towards}, we propose the Fidelity metric to evaluate the explanatory power of attribution methods and prediction models. The main idea of Fidelity is: when extracting the top-contributed feature-dimensions using an attribution method, if those extracted features have the potential to construct an optimal proxy model that agrees well with the original model on making a prediction, then this attribution method is good. In other words, we can evaluate the attribution results by measuring the consistency between the proxy model's prediction results and the original model's predictions: the more consistent these results are, the better the explanatory power of the attribution method will be. In experiments, the Fidelity metric proves that the IG method interprets better than the SG method in MOBA game event prediction tasks.

In summary, our contributions are three-fold: 
\textcolor{black}{1) We propose to predict four representative tasks (``win'', ``Tyrant'', ``kill'', and ``be-kill'') based on a large-scale \emph{HoK} dataset.}
2) We achieve interpretable event predictions with two SOTA sequence modeling networks and two SOTA gradient-based attribution methods. These works can serve as strong baselines for this task in future studies. 3) We evaluate the explanatory power of attribution methods and prediction models, proposing the Fidelity metric to quantitatively measure how well they interpret the prediction results.

\section{Related Works}
\subsection{MOBA Game Prediction}
Studies on predicting MOBA games consist of pre-game and in-game predictions. \textcolor{black}{They can also be divided into the outcome (``win'') predictions and the event predictions.} \textcolor{black}{Pre-game (outcome/``win'') predictions focus on training prediction models based on pre-game features such as hero-selections\footnote{Heroes are the characters controlled by human-players.\label{hero}} and players' historical records.} Among these works, \cite{conley2013does} is the first to predict the \emph{DotA2} results before the games start. \cite{kalyanaraman2014win} follows this work by combining the Genetic Algorithm with Logistic Regression (LR) and reports a higher prediction accuracy. \cite{kinkade2015dota} proposes to build prediction models for \emph{DotA2} based on two different sets of training data: one comprising only the hero-selection information, and the other consisting of the full post-game data. \cite{song2015predicting,semenov2016performance,makarov2017predicting} further evaluate the performances of several machine learning methods for \emph{DotA2} win predictions, including Logistic Regression, Naive Bayes, GBDT, and other methods.
\textcolor{black}{\cite{song2015predicting} also uses ``hero-lineups'' and Logistic Regression to predict the winning side of \emph{DotA2}. \cite{semenov2016performance} evaluates the performance of several machine learning methods such as Naive Bayes and Logistic Regression based on the hero-selections in \emph{Dota2}. \cite{makarov2017predicting} predicts the winning teams and their winning probabilities for \emph{DotA2} and another FPS electronic game, \emph{CSGO}.} 
\cite{wang2018outcome} proposes to improve pre-game \emph{DotA2} prediction using a better representation of hero-selection information.

Although much research has been done to perform pre-game predictions for MOBA games, in-game predictions are more informative and useful. \textcolor{black}{Therefore, recent research focuses more on in-game MOBA game outcome and event predictions.} \cite{yang2016real} first introduces three in-game features to achieve win predictions for \emph{DotA2}. \cite{hodge2019win} uses machine learning methods such as Logistic Regression and Decision Tree to predict the results of \emph{DotA2} using professional-level in-game data.
\cite{yang2020interpretable} proposes a two-stage model TSSTN to perform interpretable in-game predictions for \emph{HoK}. However, this model can attribute the results to only six human-selected features such as ``gold'' and ``heroes''. Moreover, their work achieves interpretability at the cost of accuracy, which undermines the performance. \textcolor{black}{SHAP method proposed in \cite{lundberg2017unified} can also give interpretation for the prediction procedures of the models.} Value-based reinforcement learning methods \cite{sun2018tstarbots,ye2020mastering,zhang2019hierarchical,OpenAI:OpenAI_dota} can also give in-game win predictions without interpretability. \cite{smerdov2020collection,smerdov2020detecting} try to predict the results of \emph{LoL} using multiple data collected from sensors and other hardware. \textcolor{black}{For event predictions, \cite{katona2019time} aims to predict the events inside the games such as hero deaths.} \textcolor{black}{Instead of predicting the outcome of esports matches, this work focuses on the ``micro-predictions'' in the games, in particular the predictions for the heroes' death.}

\subsection{Gradient-Based Attribution Methods}

In order to attribute the prediction of a deep network to its input features, \cite{sundararajan2017axiomatic} proposes a gradient-based attribution method \textit{Integrated Gradients} (IG). Two fundamental axioms, \textit{Sensitivity} and \textit{Implementation Invariance}, are also proposed to prove the correctness of IG. Following this work, \cite{he2019towards} applies IG to measure the word-importance for Neural Machine Translation. \cite{chen2019looks} also applies IG to interpret the results of image-recognition networks.

\cite{smilkov2017smoothgrad} proposes another gradient-based attribution method, \textit{SmoothGrad}, to identify pixels that strongly influence the final decisions of image classifiers. By adding noise to the original image, we get a set of similar images. Then by averaging the gradients for each image to the image classifiers' outputs, we can get a better sensitivity map (attribution result) of the original image to the classification result.

\subsection{Fidelity Metric For Explanatory Power}
The idea of ``fidelity'' is primarily used in the domain of Model Compression \cite{bucilua2006model,polino2018model} and Model Distillation \cite{hinton2015distilling,liu2018distilling}. Recently, \cite{jacovi2020towards} proposes to utilize a similar concept of ``faithfulness'' to evaluate the interpretable methods for deep-learning based Nature Language Processing (NLP) models. \cite{li2020evaluating} further proposes a fidelity-based metric and its practical approximation method for Neural Machine Translation (NMT).

\textcolor{black}{\subsection{Explainable Artificial Intelligence (XAI)}
With the purpose of explaining the autonomous decisions and actions of the artificial intelligence models to human users, much research effort has been devoted to finding the rationale for models' decisions. Among these researches, \cite{adadi2018peeking,gunning2019xai,dovsilovic2018explainable,samek2019towards,arrieta2020explainable,das2020opportunities,vilone2020explainable,angelov2021explainable,longo2020explainable} give thorough surveys on this topic. While many XAI methods can be applied to the esports-prediction scenarios, we choose to study the gradient-based methods as our attribution methods and their evaluation due to the progress schedule limit of our experiments. We believe further research can be done to promote this area.}

\section{Task and Dataset}
\subsection{Prediction Tasks}
We propose to predict four important events for \emph{HoK} games: ``win'', ``Tyrant\textsuperscript{\ref{tyrant}}'', ``kill'', and ``be-kill''. These are four of the most important events in \emph{HoK} and other MOBA games. The descriptions of these four events are as follows:

\begin{itemize}
    \item \textbf{Win}: Predicting which team will win the game. \textcolor{black}{The ``win'' prediction task is the most generally acknowledged and the most studied among all the tasks. Throughout all the matches, the audience and the players care most about which team will win and why.}
    \item \textbf{Tyrant\footnote{A type of Boss Monster in \emph{HoK}\textcolor{black}{, nestled into one side of the middle river}. The team that kills the Tyrant will get bonus gold and experience\textcolor{black}{, and therefore get a temporary advantage over the rival team. The “Tyrant” prediction task can be easily generalized to the prediction tasks for other boss monsters in other MOBA games, such as the ``Rift Herald'', ``Baron Nashor'', and ``Dragons'' in \emph{LoL}, and the ``Roshan'' in \emph{DotA2}.}\label{tyrant}}}: Predicting which team will seize the Tyrant. \textcolor{black}{The ``Tyrant'' task is representative of all the tasks to predict which team will kill a boss monster.}
    \item \textbf{Kill}: Predicting who will be the next killer (the hero who kills the enemy). \textcolor{black}{The event that one hero kills an enemy is one of the major focuses in MOBA games.}
    \item \textbf{Be-kill}: Predicting which hero will be killed next. \textcolor{black}{The death of a hero may greatly affect the situations and results of the games.}
\end{itemize}

\subsection{Events Extraction}
\begin{figure}[h]
\centering
\includegraphics[scale=0.5]{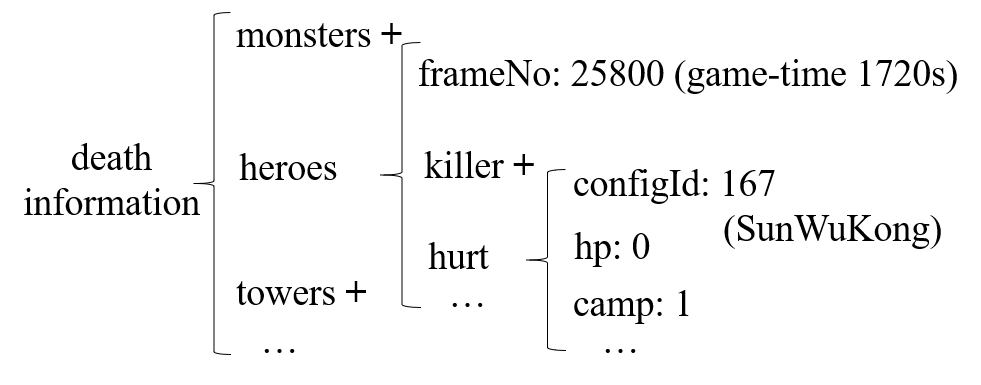}
\caption{An example of death information in the dataset.}
\label{fig:death_information}
\end{figure}

To predict the four events mentioned above, we record the death information of all the heroes, monsters, and towers of 50,278 \emph{HoK} games. \textcolor{black}{The data is collected from the top 1\% ranked (the ``Conquerors'' level) players' games happened in the first quarter of 2020.} The death information contains the death-frame (game-time), killer-information, hurt-information, and some other useful information, as shown in Figure \ref{fig:death_information}. Then the task-labels can be extracted from this death information. It is worth noting that one may extract more potential events from the death information, such as towers' destruction and the next optimal equipment.

\subsection{Feature Extraction}
In addition to the death information, we also record more than 2,000 in-game features every second of the games. We classify these features into five categories: ``hero'', ``global'', ``monster'', ``soldier'', and ``tower''. 

\begin{itemize}

    \item \textbf{Hero} \textcolor{black}{\textit{Heroes} are the game characters that are controlled by the players.} As shown in Figure \ref{fig:hero_features}, hero-features contain the information of the ten heroes in the game, including the hero's ID (name), camp, level, kill-count, assist-count, death-count, skills' information, and many other features.
    
    \begin{figure}[h]
    \centering
    \includegraphics[scale=0.5]{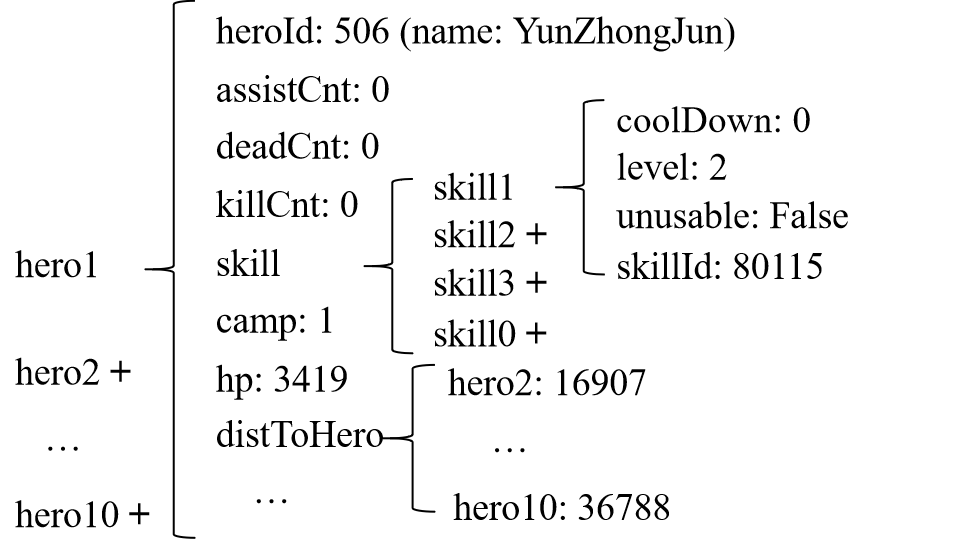}
    \caption{An example of hero-features in the dataset.}
    \label{fig:hero_features}
    \end{figure}
    
    \item \textbf{Global} Global features describe the game's overall situations, including the game-time, number of two camps' alive heroes, money-amount, and number of alive towers. An example of global features is shown in Figure \ref{fig:global_features}.
    
    \begin{figure}[h]
    \centering
    \includegraphics[scale=0.41]{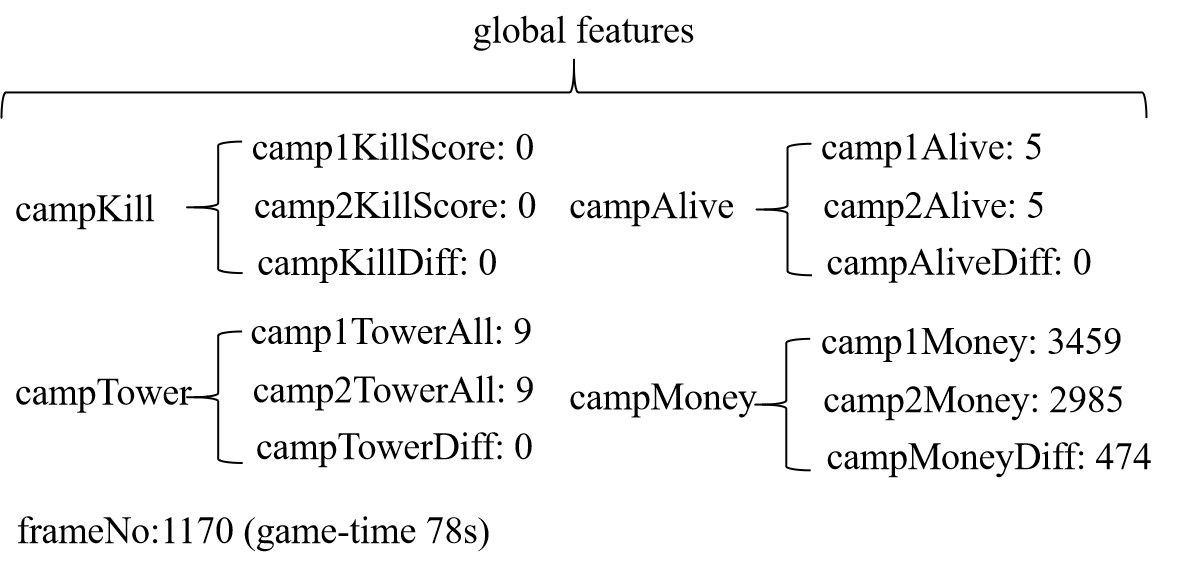}
    \caption{An example of global features in the dataset.}
    \label{fig:global_features}
    \end{figure}
    
    \item \textbf{Monster} \textcolor{black}{\textit{Monsters} refer to the neutral creatures in the Wild. By killing the monsters, heroes can obtain gold, experience, and buff (for some special monsters).} Monster-features contain the information of up to 27 monsters (Tyrant is one of them), including the monster's health-points~(hp), alive-or-not status, location, attack, and monster-type, as shown in Figure \ref{fig:monster_features}.
    \begin{figure}[h]
    \centering
    \includegraphics[scale=0.5]{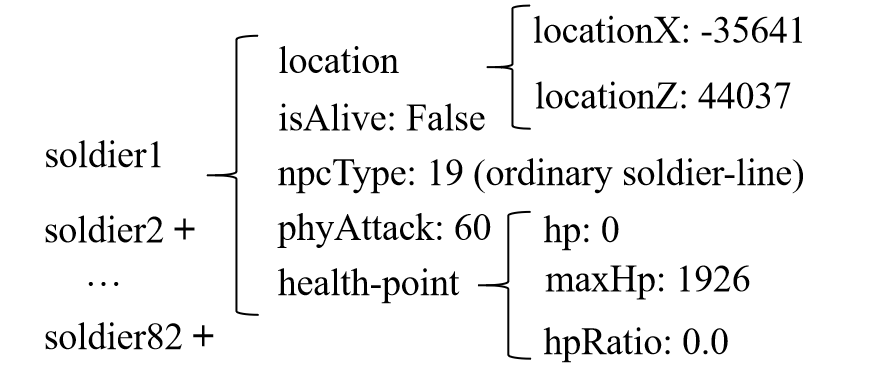}
    \caption{An example of monster-features in the dataset.}
    \label{fig:monster_features}
    \end{figure}
    
    \item \textbf{Soldier} \textcolor{black}{\textit{Soldiers} are the game units of the two teams that automatically generate and move to attack the rival heroes, tower, and base. By killing the rival soldiers, the heroes will get gold and experience.} As shown in Figure \ref{fig:soldier_features}, soldier-features cover the information of up to 82 soldiers. Features in this category include the soldiers' camp, location, hp, alive-or-not status, soldier-type, and attack.
    
    \begin{figure}[h]
    \centering
    \includegraphics[scale=0.5]{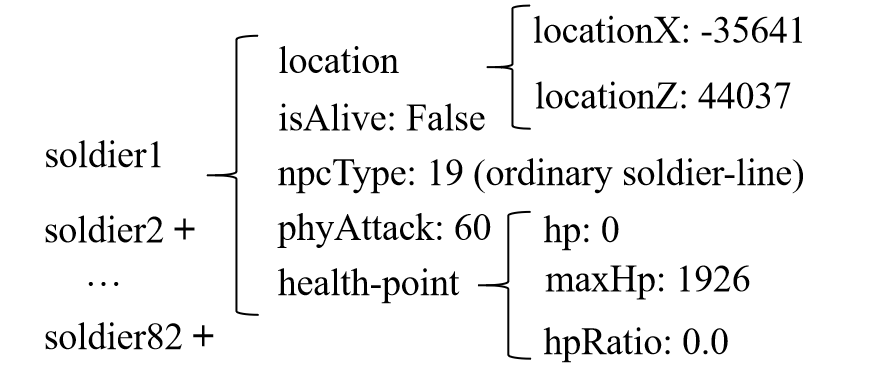}
    \caption{An example of soldier-features in the dataset.}
    \label{fig:soldier_features}
    \end{figure}
    
    \item \textbf{Tower} \textcolor{black}{\textit{Towers} refer to the defense towers of the two teams that stand in the three lanes and around the base. They would automatically attack the rival soldiers and heroes that get near them.} Tower-features contain the information of two camps' 22 towers, including the tower's attack-range, location, camp, distance to heroes, hp, tower-type, and attack. An illustration is shown in Figure \ref{fig:tower_features}.
    
    \begin{figure}[h]
    \centering
    \includegraphics[scale=0.5]{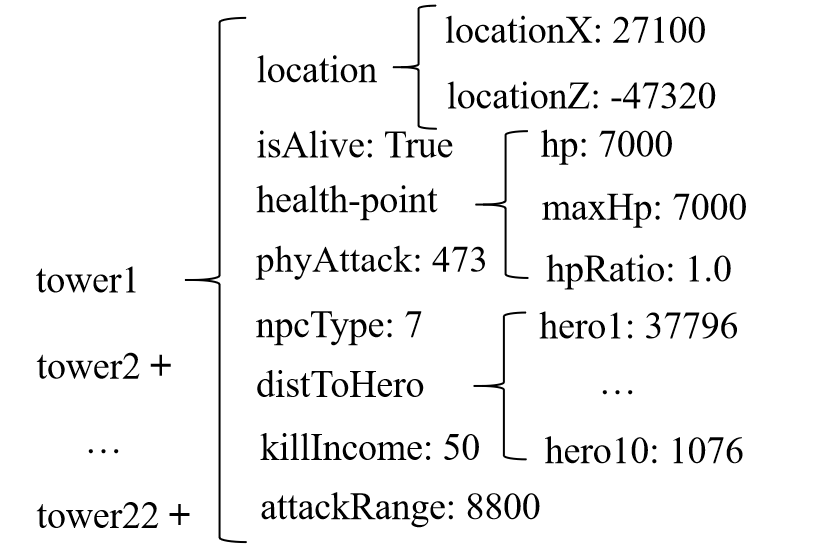}
    \caption{An example of tower-features in the dataset.}
    \label{fig:tower_features}
    \end{figure}
    
\end{itemize}

\section{Prediction}
In this section, we first encode all the categorical features in the dataset into one-hot vectors, then concatenate them with other numerical features as the input vectors. Given the encoded input feature, we train two SOTA sequence modeling networks, LSTM and Transformer, to predict the occurrences of the aforementioned events.
\begin{figure}[t]
\centering
\includegraphics[scale=0.5]{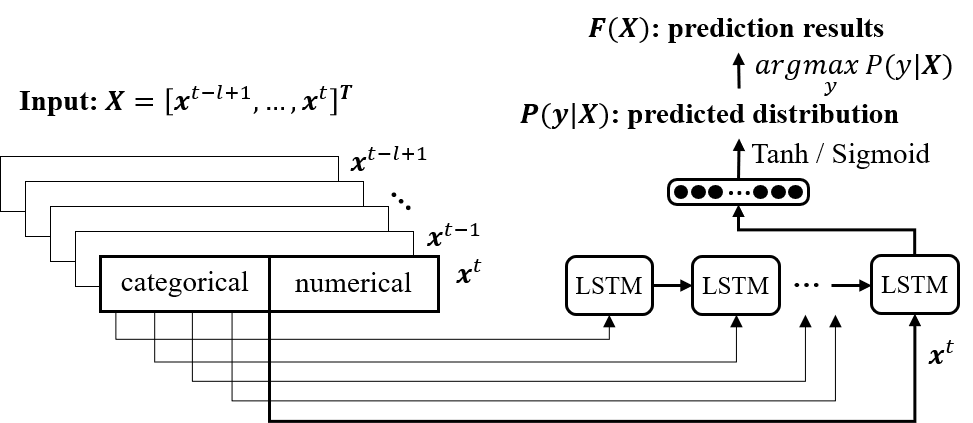}
\caption{The prediction procedure of the $l$-seconds' data and sequence modeling network LSTM for the four tasks.}
\label{fig:prediction_procedure}
\end{figure}

\subsection{Input Feature}
Some of the in-game features in the collected dataset are categorical, such as hero-ID, skill-ID, and NPC-type. To better represent these categorical features, we encode them to one-hot vectors. As for the numerical features such as gold-difference and kill-difference, we normalize them to real values ranging in $[0, 1]$. After preprocessing, all the vectors and variables will be concatenated into a 5,885-dimension vector. To capture the sequential characteristics of the input data, we choose the consecutive $l$-seconds' data as input, which means that the data up to game-time $t$ will be represented by $\bm{X} = [\bm{x}^{t-l+1}, \ldots, \bm{x}^t]^T$.  For ``Tyrant'', ``kill'', and ``be-kill'' tasks, we choose the training data at $S$ seconds-intervals before the events' happening time (using the data from game-time $t-l+1$ to $t$ to predict the event at $t + S$); for ``win'' task, we set some fixed time-intervals, and choose the game records at these intervals as the training data.

\subsection{Prediction Model}
To capture the time-sequential characteristics of the games, we use SOTA sequence modeling networks such as LSTM \cite{hochreiter1997long} and Transformer \cite{vaswani2017attention} to perform the prediction tasks. A fully-connected layer is used to make the final predictions for the four tasks, as shown in Figure \ref{fig:prediction_procedure}. \textcolor{black}{We use the same model structures for the tasks with the same input and output format, such as the ``kill'' and the ``be-kill'' tasks. For explicitness, we train a separate prediction model for each experimental setting.}

\section{Attribution Method}
In order to find the underlying reasons for the prediction models' results, we utilize two gradient-based attribution methods, \textit{Integrated Gradients} (IG) \cite{sundararajan2017axiomatic} and \textit{SmoothGrad} (SG) \cite{smilkov2017smoothgrad}, to interpret the event predictions by attributing the prediction results to the input features. Specifically, IG fulfills this task by calculating the straight-line path-integral of the gradient from a baseline input $\bm{X}'$ to the current input $\bm{X}$, while SG performs attribution by averaging the gradients of a set of similar inputs generated by adding Gaussian noise to the original input. However, there is a challenge in that the categorical features cannot be logically divided or added with noise. Therefore, to apply these attribution methods in our task, we deliberately design an additional embedding layer that maps the categorical features into continuous representations. 

\subsection{Integrated Gradients}
The \textit{Integrated Gradients} (IG) method was first proposed by \cite{sundararajan2017axiomatic} to attribute the results of deep networks to the input features. In this work, we use IG to find the top-contributing feature-dimensions for the four prediction tasks in MOBA games.

Specifically, let $\bm{x}^t = [x_1^t, \ldots, x_n^t]^T$ be the input vector at game-time $t$ and let $\bm{X} = [\bm{x}^{t-l+1}, \ldots, \bm{x}^t]^T$ be the $l$-second sequential input up to game-time $t$. Assume that $F$ is a prediction model and $P(y|\bm{X})$ is its output. We set $\bm{X}'$, which has the same dimension as $\bm{X}$, to be the baseline, with all its elements to be $0$. Then the IG of $\bm{X}$ is defined by the integral of gradient from $\bm{X}'$ to $\bm{X}$ in the straight-line path:
\begin{equation}
    IG_{i, j} = (\bm{X}_{i, j} - \bm{X}'_{i, j})\int_{\alpha = 0}^1\left.\frac{\partial P(y|\bm{\hat X})}{\partial\bm{\hat X}_{i, j}}\right|_{\bm{\hat X} = \bm{X}' + \alpha (\bm{X} - \bm{X}')}d\alpha,
\label{equation:original_ig}
\end{equation}
where $IG_{i, j}$ represents the contribution of feature-dimension $j$ to the prediction in $\bm{x}^i$.

However, this theoretical formulation of IG is inconvenient for practical applications due to the existence of the path-integral. A practical approximation of Equation (\ref{equation:original_ig}) can then be formulated by:
\begin{equation}
    IG_{i, j} \approx \frac{\bm{X}_{i, j} - \bm{X}'_{i, j}}{steps}\sum_{k=1}^{steps}\left.\frac{\partial P(y|\bm{\hat X})}{\partial\bm{\hat X}_{i, j}}\right|_{\bm{\hat X} = \bm{X}' + \frac{k}{steps} (\bm{X} - \bm{X}')},
\label{equation:approx_ig}
\end{equation}
where $steps$ is the number of steps that evenly distribute from the baseline $\bm{X}'$ to the input $\bm{X}$. The larger $steps$ we choose, the better approximation of IG we will get. In practice, $steps$ ranging from $100$ to $300$ results in good enough approximations and reasonable efficiency \cite{sundararajan2017axiomatic}.

\subsection{SmoothGrad}
\textit{SmoothGrad} is also a gradient-based attribution method, first proposed by \cite{smilkov2017smoothgrad}, which can be utilized to attribute the prediction to the input features for MOBA games. Assume that the output of the prediction network is $P(y|\bm{X})$ where $y$ is the target and $\bm{X}$ is the input. Then the SG for the $j$-th dimension of $\bm{x}^i$ in $\bm{X} = [\bm{x}^{t-l+1}, \ldots, \bm{x}^t]^T$ is calculated by:
\begin{equation}
    SG_{i, j} = \frac{1}{steps}\sum_{k=1}^{steps} \left.\frac{\partial P(y|\bm{\hat X})}{\partial \bm{\hat X}_{i, j}}\right|_{\bm{\hat X} = \bm{X} + \mathcal N (0, \sigma^2)},
\label{equation:sg}
\end{equation}
where $steps$ is the number of generated samples and $\mathcal N (0, \sigma^2)$ represents the zero-mean Gaussian noise with standard deviation $\sigma$. $SG_{i, j}$ represents the contribution of feature-dimension $j$ to the model's output $P(y|\bm{X})$ in $\bm{x}^i$.

\subsection{Embedding}

\begin{figure}[t]
\centering
\includegraphics[scale=0.70]{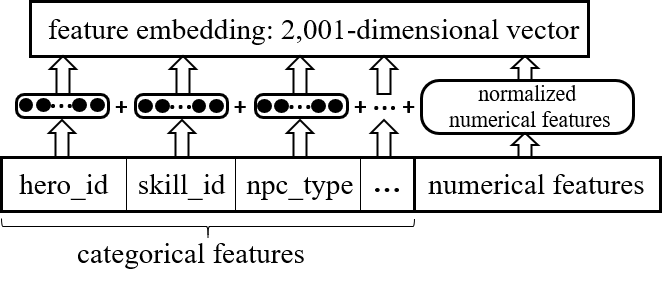}
\caption{The embedding procedure to convert the 5,885-dimension input into 2,001-dimension embedding vectors.}
\label{fig:embedding_procedure}
\end{figure}

To make the categorical features in the input continuous, we need to transform the original input into embedding vectors first. As Figure \ref{fig:embedding_procedure} shows, the input vector of 5,885 dimensions will first be mapped into embedding vectors of 2,001 dimensions, in which numerical features will be directly copied after normalization and categorical features such as hero-ID and skill-ID will be transformed using several parallel fully-connected (FC) layers. Input dimensions belonging to the same feature will be processed by the same FC layer for the purpose of attribution. 

\section{Fidelity Metric}

                

\renewcommand{\algorithmicrequire}{\textbf{Input:}}  
\renewcommand{\algorithmicensure}{\textbf{Output:}} 
\begin{algorithm}[t]
    \fontsize{11pt}{11pt}\selectfont
        \caption{Fidelity}
        \begin{algorithmic}[1] 
                \Require 
                Original prediction model $F$;
                Proxy model ${Q}$;
                Attribution method ${\phi}$;
                (Hyper-parameter) Number of selected top-contributed feature-dimensions $k$;
                Training set $Tr$;
                Testing set $T$
                
                \Ensure $Fidelity$;
                \State ${V_{\phi}^k} \gets \{\}$;
                ${W_{\phi}^k} \gets \{\}$;
                \For {$\bm{X}$ in $Tr$ and $T$}
                \State $\overline{\bm{X}} \gets$ Preserving the top-$k$ contributed feature-dimensions of $\bm{X}$ using $\phi$ and $F$, and zero-masking the rest dimensions
                \If {$\bm{X}\in Tr$}
                \State ${V_{\phi}^k} \gets {V_{\phi}^k} \cup \{\overline{\bm{X}}; F(\bm{X})\}$
                \Else  
                \State  ${W_{\phi}^k} \gets {W_{\phi}^k} \cup \{\overline{\bm{X}}; F(\bm{X})\}$
                \EndIf
                \EndFor
                \State Train $Q$ using ${V_{\phi}^k}$
                \State  
                $Fidelity = Accuracy[Q(\bm{X}) = y | \bm{X}, y \in W_{\phi}^k]$
        \end{algorithmic}  
        \label{alg:fidelity}
\end{algorithm}

``Fidelity'' is the concept of keeping part of the input and assessing how much information can be retrieved, which is recently formulated to evaluate the explanation methods in NLP \cite{li2020evaluating, jacovi2020towards}. In this section, we propose a fidelity-based metric that measures the explanatory power of models and attribution methods in our MOBA game prediction tasks. As shown in Algorithm \ref{alg:fidelity}, with attribution method $\phi$, we can attribute the prediction results of model $F$ to $k$ top-contributed feature-dimensions. Then we attempt to evaluate how well these attributed features represent the input.  Specifically, we only preserve the top-$k$ contributed feature-dimensions for each training or testing instance and replace other dimensions with zero. In this way, we get new sets of training and testing dataset $V_{\phi}^k$ and $W_{\phi}^k$ from the original sets $Tr$ and $T$. Then, an optimal proxy model $Q$, which has the same architecture with $F$, can be trained using the new training set $V_{\phi}^k$. Finally, the Fidelity of attribution method $\phi$ on model $F$ can be computed as follows:

\begin{equation}
    Fidelity_{Tr,T}^k(F, \phi) = Accuracy\textbf{[}Q(\bm{X}) = y | \bm{X}, y \in W_{\phi}^k\textbf{]},
\label{equation:original_fidelity}
\end{equation}
in which $F$ is the original prediction model, $Q$ is the proxy model, $\phi$ is the attribution method, $Tr$ and $T$ are the original training and testing sets upon which $F$ is trained and tested, and $W_{\phi}^k$ represents the new testing set containing instances that keep the top-$k$ contributed feature-dimensions selected by $\phi$ for instances of $T$.

\section{Experiments}
\subsection{Experimental Settings}
We conduct experiments on the dataset mentioned above. We set 5,000 games from the dataset as the validation set and another 5,000 games as the testing set. The rest 40,278 games are the training set.

The sequence-length $l$ of the sequential input $\bm{X} = [\bm{x}^{t-l+1}, \ldots, \bm{x}^t]^T$ is $5$.\footnote{\textcolor{black}{Prior experiments showed that using data with sequence length longer than 5 did not bring evident improvement to the accuracy.}} \textcolor{black}{For ``Tyrant'', ``kill'', and ``be-kill'' tasks, we choose the input data at $S$ seconds-intervals before the events' happening time, where $S \in \{1, 5, 10, 15, 20\}$.} For ``win'' task, we choose the input data every $60$ seconds from the beginning of each game. We also assess the accuracy of the ``win'' task at different game-times, ranging from 40.0-seconds (the beginning of the games) to 20.0-minutes. \textcolor{black}{The accuracy of the ``win'' tasks is accessed averagely (with respect to the predictions at different game-times) and separately.}

\subsection{Prediction Models}

\subsubsection{LSTM} We use bidirectional LSTM with two recurrent layers. The probability of dropout is $0.2$. The size of the hidden state is $128$. After the LSTM, we use a $256$-dimension fully-connected layer and a $tanh$ function to compute the class-scores $P(y|\bm{X})$.

\subsubsection{Transformer} The numbers of layers and attention heads are $2$ and $8$, respectively. We set the dropout probability of the Transformer to be $0.1$ and the hidden dimension to be $256$. After the Transformer, a $256$-dimension FC layer and $tanh$ function are used to compute $P(y|\bm{X})$.

\subsection{Attribution Methods}
We compare the ``explanatory power'' of two attribution methods, \textit{Integrated Gradients} (IG) and \textit{SmoothGrad} (SG), on two prediction models (LSTM and Transformer) and four tasks (``Tyrant'', ``win'', ``kill'', and ``be-kill''). For each task, we use the attribution methods to find the top-$k$ contributed feature-dimensions in the input, where $k \in \{100, 10, 5, 1\}$. Specifically, since our input is time-sequential, we average the IG/SG of the input among the time-dimension of the input $\bm{X}$, then choose the top-$k$ dimensions of the averaged IG/SG to be the top-contributed feature-dimensions.

\subsubsection{Integrated Gradients} We apply Equation (\ref{equation:approx_ig}) to realize IG, and choose the dividing steps to be $steps = 100$.

\subsubsection{SmoothGrad} We apply Equation (\ref{equation:sg}) to realize SG with $steps = 100$, and set the standard derivation $\sigma$ of Gaussian noise for the $i$-th dimension of $\bm{X}$ to be $0.15 \cdot (max(\bm{X}_i) - min(\bm{X}_i))$.\footnote{We do not fine-tune this parameter too much because the result is not sensitive to the value of $\sigma$.}

\subsection{Evaluation Metrics} \subsubsection{Accuracy} We use the prediction accuracy as the evaluation metric for the aforementioned two prediction models for the four tasks.
\subsubsection{Fidelity} We evaluate the explanatory power of different pairs of attribution method (IG and SG) and prediction model (LSTM and Transformer) with the Fidelity metric using Equation (\ref{equation:original_fidelity}) and Algorithm \ref{alg:fidelity}. We conduct these experiments by preserving the top-$k$ contributed feature-dimensions, where $k \in \{100, 10, 5, 1\}$.


\begin{table}[t]
\caption{Accuracy of LSTM and Transformer for the four tasks \textcolor{black}{averaged across the game-times or} at different intervals before the events' happening.}
\resizebox{87mm}{!}{
\begin{tabular}{|c|c|ccccc|}
\hline
\multirow{2}{*}{\textbf{Task}} & \multirow{2}{*}{\textbf{Model}} & \multicolumn{5}{c|}{\textbf{Accuracy}} \\ \cline{3-7} 
 & & \textcolor{black}{1.0sec} & 5.0sec & 10.0sec & 15.0sec & 20.0sec \\ \hline
\multirow{3}{*}{Tyrant} & LSTM & \textcolor{black}{0.939} & 0.930 & 0.910 & 0.871 & 0.834 \\
 & Transformer & \textcolor{black}{\textbf{0.944}} & \textbf{0.934} & \textbf{0.916} & \textbf{0.880} & \textbf{0.841} \\ \hline
\multirow{3}{*}{win} & LSTM & \multicolumn{5}{c|}{0.704} \\
 & Transformer & \multicolumn{5}{c|}{\textbf{0.708}} \\ \hline
\multirow{3}{*}{kill} & LSTM & \textcolor{black}{0.274} & 0.221 & 0.206 & 0.184 & 0.175 \\
 & Transformer & \textcolor{black}{\textbf{0.281}} & \textbf{0.230} & \textbf{0.216} & \textbf{0.189} & \textbf{0.179} \\ \hline
\multirow{3}{*}{be-kill} & LSTM & \textcolor{black}{0.357} & 0.278 & 0.207 & 0.160 & 0.143  \\
 & Transformer & \textcolor{black}{\textbf{0.370}} & \textbf{0.319} & \textbf{0.228} & \textbf{0.175} & \textbf{0.149} \\ \hline
\end{tabular}
}
\label{tab:accuracy}
\end{table}

\section{Results and Analysis}
\subsection{Prediction Accuracy}


\begin{figure}[t]
\centering
\includegraphics[scale=0.48]{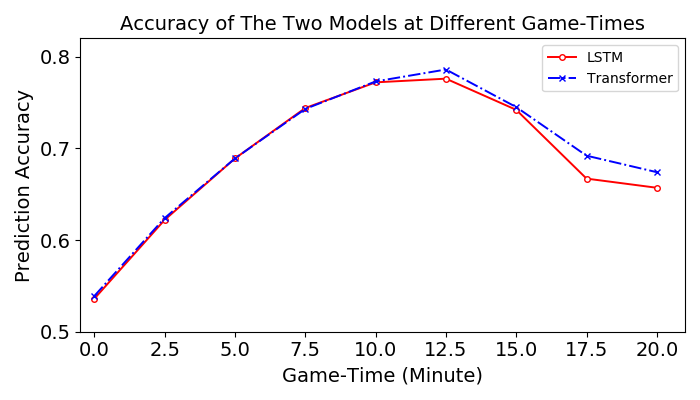}
\caption{The ``win'' prediction accuracy of the two prediction models at different game-times ranging from 40.0-seconds (the beginnings of the games) to 20.0-minutes the late game-stages.}
\label{fig:win_time}
\end{figure}

Table \ref{tab:accuracy} indicates that the two models achieve close accuracy, while Transformer is slightly more accurate. \textcolor{black}{For ``Tyrant'', ``kill'', and ``be-kill'' tasks, the prediction accuracy decreases when the prediction interval $S$ increases (from 1.0-seconds to 20.0-seconds), which is logical since it is easier to predict an event in the near future than the one in the distant future.} For example, if the Tyrant is killed at game-time $t$-seconds, then at ``$t-5$''-seconds we are almost sure the team that has an advantage will seize the Tyrant, while at ``$t-20$''-seconds the future situations are not that clear. \textcolor{black}{For the ``win'' tasks, Table \ref{tab:accuracy} shows the average prediction accuracy across the different game-times.}

From Table \ref{tab:accuracy}, we can also conclude that ``Tyrant'' prediction is the most accurate one and ``win'' prediction is the next, while ``kill'' and ``be-kill'' predictions are less accurate. The underlying reasons are as follows: 1) ``Tyrant'' and ``win'' are binary-classification tasks, while there are ten labels for ``kill'' and ``be-kill''. 2) It is much easier to predict the macro-scale events (such as which team will seize the Tyrant and which team will win) than to predict the micro-scale events (which hero exactly will be killed or kill others) since there is too much uncertainty and variability for micro-scale events. 
\textcolor{black}{From Table \ref{tab:accuracy}, we can also find that as ten-label classification tasks, for $S = 1$, $S = 5$, and $S = 10$, the ``be-kill'' tasks are more accurate than the ``kill'' tasks with the same experimental settings, while for $S = 15$ and $S = 20$ seconds, the ``be-kill'' tasks are less accurate.
This means that predictions for ``be-kill'' tasks are more accurate in the near future (within 10 seconds), while ``kill'' predictions are more accurate in the distant future (longer than 10 seconds).} A possible explanation for this phenomenon is as follows: In the near future, it is easier to predict who will be killed by checking whose situation is the worst, while there might be several possible candidate killers, making it relatively harder to predict which one of them exactly will be the killer. However, situations will definitely change in the distant future, such as 20 seconds later. Therefore, we are no longer sure which hero will be in the worst situation then, while it is relatively more accurate to predict which hero will be the potential killer by checking who is the most powerful one.

We further test the accuracy of the two models at different game-times for the ``win'' task to investigate the nature of outcomes for MOBA games. As shown in Figure \ref{fig:win_time}, the prediction accuracy of both models first increases as the games progress, then declines in the late-game stages (after 12.5-minutes of game-play). This phenomenon happens due to the following reasons: 1) In the early-game stages (before 12.5-minutes), the games become more predictable as time goes on because the leading team will accumulate its advantages in gold, level, and equipment. 2) In the late-game stages (after 12.5-minutes), the level and equipment of both teams reach a maximum. Therefore, games are increasingly affected by uncertainty, such as players' accidental mistakes, and therefore are harder to predict.

\begin{table}[t]
\caption{Fidelity of different pairs of attribution methods and prediction models for the four prediction tasks.}
\resizebox{85mm}{!}{
\begin{tabular}{|c|c|cccc|}
\hline
\multirow{2}{*}{\textbf{Task}} & \multirow{2}{*}{\textbf{\begin{tabular}[c]{@{}c@{}}Attribution\\ + Model\end{tabular}}} & \multicolumn{4}{c|}{\textbf{Fidelity}} \\ \cline{3-6} 
 &  & Top 100 & Top 10 & Top 5 & Top 1 \\ \hline
\multirow{4}{*}{\begin{tabular}[c]{@{}c@{}}Tyrant\\ 5.0sec\end{tabular}} & IG+LSTM & 0.942 & 0.862 & \textbf{0.822} & \textbf{0.726} \\
 & SG+LSTM & 0.956 & 0.816 & 0.778 & 0.653\\
 & IG+Transformer & 0.968 & \textbf{0.885} & 0.811 & 0.611 \\
 & SG+Transformer & \textbf{0.970} & 0.840 & 0.800 & 0.632 \\ \hline
\multirow{4}{*}{win} & IG+LSTM & 0.908 & 0.814 & 0.799 & 0.715\\
 & SG+LSTM & 0.897 & 0.825 & 0.762 & 0.656 \\
 & IG+Transformer & \textbf{0.950} & \textbf{0.891} & \textbf{0.866} & \textbf{0.800} \\
 & SG+Transformer & 0.938 & 0.872 & 0.834 & 0.622 \\ \hline
\multirow{4}{*}{\begin{tabular}[c]{@{}c@{}}kill\\ 5.0sec\end{tabular}} & IG+LSTM & 0.329 & 0.207 & 0.201 & 0.161 \\
 & SG+LSTM &0.311 & 0.218 & 0.189 & 0.178 \\
 & IG+Transformer & \textbf{0.460} & \textbf{0.357} & \textbf{0.313} & \textbf{0.184} \\
 & SG+Transformer & 0.384 & 0.175 & 0.159 & 0.177 \\ \hline
\multirow{4}{*}{\begin{tabular}[c]{@{}c@{}}bekill\\ 5.0sec\end{tabular}} & IG+LSTM & 0.293 & 0.177 &  0.157 &  0.148 \\
 & SG+LSTM &  0.296 &  0.171 & 0.172 & 0.157 \\
 & IG+Transformer & 0.346 & \textbf{0.268} & \textbf{0.259} & \textbf{0.193} \\
 & SG+Transformer & \textbf{0.363} & 0.244 & 0.244 & 0.179 \\ \hline
\end{tabular}
}
\label{tab:fidelity}
\end{table}

\subsection{Fidelity}


Fidelity of different attribution methods (IG and SG) and prediction models (LSTM and Transformer) with respect to $k$ top-contributed feature-dimensions ($k \in \{100, 10, 5, 1\}$) is shown in Table \ref{tab:fidelity}. With a few exceptions, IG and Transformer achieve the highest Fidelity for ``win'', ``kill'', and ``be-kill'' tasks. Fidelity decreases when the number of top-contributed feature-dimensions $k$ changes from $100$ to $1$, because when we preserve fewer features, less information of the game can be retrieved.

Experiments show that the Fidelity for ``Tyrant'' and ``win'' tasks is higher than the Fidelity for ``kill'' and ``be-kill'' tasks. One reason is that the first two tasks are binary classification tasks and the last two tasks are ten-label classification tasks. Therefore, the Fidelity (defined by the accuracy of the proxy model) of ``Tyrant'' and ``win'' is higher. The other reason is that to predict which team will win or seize the Tyrant, we mainly rely on a small number of critical features (such as gold-difference and Tyrant's distances to heroes); however, to predict which hero will be the next killer or be-killed one, we need to consider more factors, such as hero-skill information, hero-level, locations, hp, and many other important features. 

\subsection{Parameters}
We further conduct an additional experiment to evaluate the effect of the choice of $steps$ on the final Fidelity result. We assess the Fidelity of different $steps$ values of IG and SG for the ``win'' prediction task with Transformer as the prediction model. From Table \ref{tab:fidelity_steps}, we can see that there is little change in Fidelity for $steps$ ranging in $[10, 500]$, which indicates that Fidelity is not sensitive to the choice of $steps$. 

\begin{table}[t]
\caption{Fidelity of different $steps$ values for the ``win'' prediction tasks of IG and SG methods and Transformer.}
\resizebox{85mm}{!}{
\begin{tabular}{|c|c|ccccc|}
\hline
\multirow{2}{*}{\textbf{\begin{tabular}[c]{@{}c@{}}Attribution\\ + Model\end{tabular}}} & \multirow{2}{*}{\textbf{Top}} & \multicolumn{5}{c|}{\textbf{Fidelity}} \\ \cline{3-7} 
 &  & steps10 & steps50 & steps100 & steps300 & steps500 \\ \hline
\multirow{4}{*}{\begin{tabular}[c]{@{}c@{}}IG +\\ Transformer\end{tabular}} & 100 &0.960  &0.952 &0.956 &0.954 &0.951 \\
 & 10 &0.877 &0.889 &0.890 &0.883 &0.881 \\
 & 5 &0.852 &0.867 &0.866 &0.863 &0.870 \\
 & 1 &0.792  &0.805 &0.799 &0.796 &0.792 \\ \hline
\multirow{4}{*}{\begin{tabular}[c]{@{}c@{}}SG +\\ Transformer\end{tabular}} & 100 &0.925 &0.930 &0.936 &0.933 &0.930  \\
 & 10 &0.875 &0.860 &0.870 &0.884  &0.875 \\
 & 5 &0.834 &0.838  &0.837 &0.837 &0.832 \\
 & 1 &0.622 &0.622 &0.622  &0.622 &0.622 \\ \hline
\end{tabular}
}
\label{tab:fidelity_steps}
\end{table}

\subsection{Case Study}

\begin{figure}[t]
\centering
\includegraphics[scale=0.30]{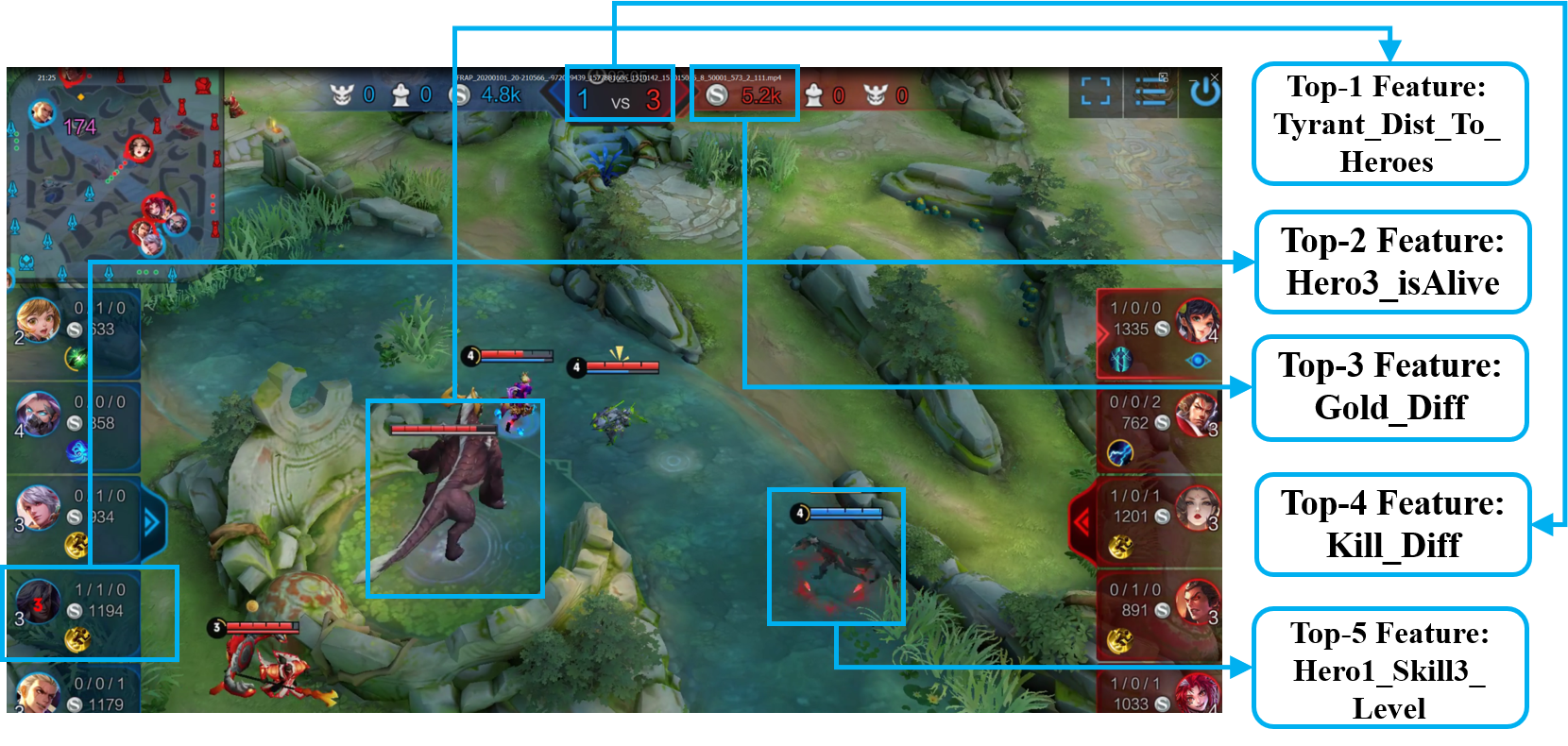}
\caption{The top-5 features attributed by IG and Transformer when two teams are fighting for Tyrant. Transformer predicts that the red team will seize the Tyrant.}
\label{fig:case_study}
\end{figure}

A case study is given in which the two teams are fighting for the Tyrant. As Figure \ref{fig:case_study} shows, Transformer predicts that the red team will get the Tyrant with probability $86$\%, and IG further attributes the prediction result to five reasons: 1)  Distances between the Tyrant and the heroes: the red team's heroes are closer to the Tyrant and therefore have a better chance of killing the Tyrant. 2) Hero-3 has died: hero-3 belongs to the blue team, so the blue team has a disadvantage. 3) Gold difference and 4) Kill-count difference: the blue team has a disadvantage in terms of gold difference and kill-count difference. 5) The skill-3 (ultimate skill) of hero-1 is of a low level: this skill is essential for group fighting, so hero-1's team (blue team) is not capable of seizing the Tyrant. Eventually, the red team indeed kills the Tyrant.

\section{Conclusions}
In this paper, we attempt to address two main issues for in-game MOBA events: 1) insufficient in-game features and 2) lack of interpretability. We first collect a large-scale \emph{HoK} dataset containing rich in-game features. To predict four important events (``Tyrant'', ``win'', ``kill'', and ``be-kill'') of \emph{HoK} in an interpretable manner, we train two sequence modeling networks (LSTM and Transformer) based on the collected dataset and adopt two attribution methods, \textit{Integrated Gradients} and \textit{SmoothGrad}, to give human-interpretable explanations of the prediction results. In addition, a fidelity-based metric is proposed to evaluate the explanatory power of the attribution methods and prediction models. Experiments show that LSTM and Transformer suit well the prediction tasks in terms of accuracy, and \textit{Integrated Gradients} outperforms \textit{SmoothGrad} in terms of the Fidelity metric in our scenarios. \textcolor{black}{With the popularity of MOBA esports, the interpretable event predictions are becoming useful for more and more MOBA related applications.} \textcolor{black}{Our research can act as a first step that could, in our future work, be followed by the development of a user facing tool and a user study that demonstrates and confirms the interpretability.}






%




\bibliographystyle{IEEEtran}
\bibliography{abc}

\begin{thebibliography}{10}
\providecommand{\url}[1]{#1}
\csname url@samestyle\endcsname
\providecommand{\newblock}{\relax}
\providecommand{\bibinfo}[2]{#2}
\providecommand{\BIBentrySTDinterwordspacing}{\spaceskip=0pt\relax}
\providecommand{\BIBentryALTinterwordstretchfactor}{4}
\providecommand{\BIBentryALTinterwordspacing}{\spaceskip=\fontdimen2\font plus
\BIBentryALTinterwordstretchfactor\fontdimen3\font minus
  \fontdimen4\font\relax}
\providecommand{\BIBforeignlanguage}[2]{{%
\expandafter\ifx\csname l@#1\endcsname\relax
\typeout{** WARNING: IEEEtran.bst: No hyphenation pattern has been}%
\typeout{** loaded for the language `#1'. Using the pattern for}%
\typeout{** the default language instead.}%
\else
\language=\csname l@#1\endcsname
\fi
#2}}
\providecommand{\BIBdecl}{\relax}
\BIBdecl

\bibitem{conley2013does}
K.~Conley and D.~Perry, ``How does he saw me? a recommendation engine for
  picking heroes in dota 2,'' \emph{Np, nd Web}, vol.~7, 2013.

\bibitem{yang2016real}
Y.~Yang, T.~Qin, and Y.-H. Lei, ``Real-time esports match result prediction,''
  \emph{arXiv preprint arXiv:1701.03162}, 2016.

\bibitem{semenov2016performance}
A.~Semenov, P.~Romov, S.~Korolev, D.~Yashkov, and K.~Neklyudov, ``Performance
  of machine learning algorithms in predicting game outcome from drafts in dota
  2,'' in \emph{International Conference on Analysis of Images, Social Networks
  and Texts}.\hskip 1em plus 0.5em minus 0.4em\relax Springer, 2016, pp.
  26--37.

\bibitem{hochreiter1997long}
S.~Hochreiter and J.~Schmidhuber, ``Long short-term memory,'' \emph{Neural
  computation}, vol.~9, no.~8, pp. 1735--1780, 1997.

\bibitem{vaswani2017attention}
A.~Vaswani, N.~Shazeer, N.~Parmar, J.~Uszkoreit, L.~Jones, A.~N. Gomez,
  {\L}.~Kaiser, and I.~Polosukhin, ``Attention is all you need,'' in
  \emph{Advances in neural information processing systems}, 2017, pp.
  5998--6008.

\bibitem{sundararajan2017axiomatic}
M.~Sundararajan, A.~Taly, and Q.~Yan, ``Axiomatic attribution for deep
  networks,'' in \emph{International conference on machine learning}.\hskip 1em
  plus 0.5em minus 0.4em\relax PMLR, 2017, pp. 3319--3328.

\bibitem{smilkov2017smoothgrad}
D.~Smilkov, N.~Thorat, B.~Kim, F.~Vi{\'e}gas, and M.~Wattenberg, ``Smoothgrad:
  removing noise by adding noise,'' \emph{arXiv preprint arXiv:1706.03825},
  2017.

\bibitem{li2020evaluating}
J.~Li, L.~Liu, H.~Li, G.~Li, G.~Huang, and S.~Shi, ``Evaluating explanation
  methods for neural machine translation,'' in \emph{Proceedings of the 58th
  Annual Meeting of the Association for Computational Linguistics}, 2020, pp.
  365--375.

\bibitem{jacovi2020towards}
A.~Jacovi and Y.~Goldberg, ``Towards faithfully interpretable nlp systems: How
  should we define and evaluate faithfulness?'' in \emph{Proceedings of the
  58th Annual Meeting of the Association for Computational Linguistics}, 2020,
  pp. 4198--4205.

\bibitem{kalyanaraman2014win}
K.~Kalyanaraman, ``To win or not to win? a prediction model to determine the
  outcome of a dota2 match,'' Technical report, University of California San
  Diego, Tech. Rep., 2014.

\bibitem{kinkade2015dota}
N.~Kinkade, L.~Jolla, and K.~Lim, ``Dota 2 win prediction,'' \emph{Univ Calif},
  vol.~1, pp. 1--13, 2015.

\bibitem{song2015predicting}
K.~Song, T.~Zhang, and C.~Ma, ``Predicting the winning side of dota2,''
  \emph{Sl: sn}, 2015.

\bibitem{makarov2017predicting}
I.~Makarov, D.~Savostyanov, B.~Litvyakov, and D.~I. Ignatov, ``Predicting
  winning team and probabilistic ratings in “dota 2” and “counter-strike:
  Global offensive” video games,'' in \emph{International Conference on
  Analysis of Images, Social Networks and Texts}.\hskip 1em plus 0.5em minus
  0.4em\relax Springer, 2017, pp. 183--196.

\bibitem{wang2018outcome}
N.~Wang, L.~Li, L.~Xiao, G.~Yang, and Y.~Zhou, ``Outcome prediction of dota2
  using machine learning methods,'' in \emph{Proceedings of 2018 International
  Conference on Mathematics and Artificial Intelligence}, 2018, pp. 61--67.

\bibitem{hodge2019win}
V.~J. Hodge, S.~M. Devlin, N.~J. Sephton, F.~O. Block, P.~I. Cowling, and
  A.~Drachen, ``Win prediction in multi-player esports: Live professional match
  prediction,'' \emph{IEEE Transactions on Games}, 2019.

\bibitem{yang2020interpretable}
Z.~Yang, Z.~Pan, Y.~Wang, D.~Cai, S.~Shi, S.-L. Huang, W.~Bi, and X.~Liu,
  ``Interpretable real-time win prediction for honor of kings--a popular mobile
  moba esport,'' \emph{IEEE Transactions on Games}, 2022.

\bibitem{lundberg2017unified}
S.~M. Lundberg and S.-I. Lee, ``A unified approach to interpreting model
  predictions,'' in \emph{Proceedings of the 31st international conference on
  neural information processing systems}, 2017, pp. 4768--4777.

\bibitem{sun2018tstarbots}
P.~Sun, X.~Sun, L.~Han, J.~Xiong, Q.~Wang, B.~Li, Y.~Zheng, J.~Liu, Y.~Liu,
  H.~Liu \emph{et~al.}, ``Tstarbots: Defeating the cheating level builtin ai in
  starcraft ii in the full game,'' \emph{arXiv preprint arXiv:1809.07193},
  2018.

\bibitem{ye2020mastering}
D.~Ye, Z.~Liu, M.~Sun, B.~Shi, P.~Zhao, H.~Wu, H.~Yu, S.~Yang, X.~Wu, Q.~Guo
  \emph{et~al.}, ``Mastering complex control in moba games with deep
  reinforcement learning.'' in \emph{AAAI}, 2020, pp. 6672--6679.

\bibitem{zhang2019hierarchical}
Z.~Zhang, H.~Li, L.~Zhang, T.~Zheng, T.~Zhang, X.~Hao, X.~Chen, M.~Chen,
  F.~Xiao, and W.~Zhou, ``Hierarchical reinforcement learning for multi-agent
  moba game,'' \emph{arXiv preprint arXiv:1901.08004}, 2019.

\bibitem{OpenAI:OpenAI_dota}
OpenAI, ``Openai five,'' \url{https://blog.openai.com/openai-five/}, 2018.

\bibitem{smerdov2020collection}
A.~Smerdov, B.~Zhou, P.~Lukowicz, and A.~Somov, ``Collection and validation of
  psycophysiological data from professional and amateur players: a multimodal
  esports dataset,'' \emph{arXiv preprint arXiv:2011.00958}, 2020.

\bibitem{smerdov2020detecting}
A.~Smerdov, A.~Somov, E.~Burnaev, B.~Zhou, and P.~Lukowicz, ``Detecting video
  game player burnout with the use of sensor data and machine learning,''
  \emph{IEEE Internet of Things Journal}, vol.~8, no.~22, pp. 16\,680--16\,691,
  2021.

\bibitem{katona2019time}
A.~Katona, R.~Spick, V.~J. Hodge, S.~Demediuk, F.~Block, A.~Drachen, and J.~A.
  Walker, ``Time to die: Death prediction in dota 2 using deep learning,'' in
  \emph{2019 IEEE Conference on Games (CoG)}.\hskip 1em plus 0.5em minus
  0.4em\relax IEEE, 2019, pp. 1--8.

\bibitem{he2019towards}
S.~He, Z.~Tu, X.~Wang, L.~Wang, M.~Lyu, and S.~Shi, ``Towards understanding
  neural machine translation with word importance,'' in \emph{Proceedings of
  the 2019 Conference on Empirical Methods in Natural Language Processing and
  the 9th International Joint Conference on Natural Language Processing
  (EMNLP-IJCNLP)}, 2019, pp. 953--962.

\bibitem{chen2019looks}
C.~Chen, O.~Li, D.~Tao, A.~Barnett, C.~Rudin, and J.~K. Su, ``This looks like
  that: deep learning for interpretable image recognition,'' in \emph{Advances
  in neural information processing systems}, 2019, pp. 8930--8941.

\bibitem{bucilua2006model}
C.~Buciluǎ, R.~Caruana, and A.~Niculescu-Mizil, ``Model compression,'' in
  \emph{Proceedings of the 12th ACM SIGKDD international conference on
  Knowledge discovery and data mining}, 2006, pp. 535--541.

\bibitem{polino2018model}
A.~Polino, R.~Pascanu, and D.~Alistarh, ``Model compression via distillation
  and quantization,'' \emph{arXiv preprint arXiv:1802.05668}, 2018.

\bibitem{hinton2015distilling}
G.~Hinton, O.~Vinyals, and J.~Dean, ``Distilling the knowledge in a neural
  network,'' \emph{stat}, vol. 1050, p.~9, 2015.

\bibitem{liu2018distilling}
Y.~Liu, W.~Che, H.~Zhao, B.~Qin, and T.~Liu, ``Distilling knowledge for
  search-based structured prediction,'' in \emph{Proceedings of the 56th Annual
  Meeting of the Association for Computational Linguistics (Volume 1: Long
  Papers)}, 2018, pp. 1393--1402.

\bibitem{adadi2018peeking}
A.~Adadi and M.~Berrada, ``Peeking inside the black-box: a survey on
  explainable artificial intelligence (xai),'' \emph{IEEE access}, vol.~6, pp.
  52\,138--52\,160, 2018.

\bibitem{gunning2019xai}
D.~Gunning, M.~Stefik, J.~Choi, T.~Miller, S.~Stumpf, and G.-Z. Yang,
  ``Xai—explainable artificial intelligence,'' \emph{Science Robotics},
  vol.~4, no.~37, 2019.

\bibitem{dovsilovic2018explainable}
F.~K. Do{\v{s}}ilovi{\'c}, M.~Br{\v{c}}i{\'c}, and N.~Hlupi{\'c}, ``Explainable
  artificial intelligence: A survey,'' in \emph{2018 41st International
  convention on information and communication technology, electronics and
  microelectronics (MIPRO)}.\hskip 1em plus 0.5em minus 0.4em\relax IEEE, 2018,
  pp. 0210--0215.

\bibitem{samek2019towards}
W.~Samek and K.-R. M{\"u}ller, ``Towards explainable artificial intelligence,''
  in \emph{Explainable AI: interpreting, explaining and visualizing deep
  learning}.\hskip 1em plus 0.5em minus 0.4em\relax Springer, 2019, pp. 5--22.

\bibitem{arrieta2020explainable}
A.~B. Arrieta, N.~D{\'\i}az-Rodr{\'\i}guez, J.~Del~Ser, A.~Bennetot, S.~Tabik,
  A.~Barbado, S.~Garc{\'\i}a, S.~Gil-L{\'o}pez, D.~Molina, R.~Benjamins
  \emph{et~al.}, ``Explainable artificial intelligence (xai): Concepts,
  taxonomies, opportunities and challenges toward responsible ai,''
  \emph{Information fusion}, vol.~58, pp. 82--115, 2020.

\bibitem{das2020opportunities}
A.~Das and P.~Rad, ``Opportunities and challenges in explainable artificial
  intelligence (xai): A survey,'' \emph{arXiv preprint arXiv:2006.11371}, 2020.

\bibitem{vilone2020explainable}
G.~Vilone and L.~Longo, ``Explainable artificial intelligence: a systematic
  review,'' \emph{arXiv preprint arXiv:2006.00093}, 2020.

\bibitem{angelov2021explainable}
P.~P. Angelov, E.~A. Soares, R.~Jiang, N.~I. Arnold, and P.~M. Atkinson,
  ``Explainable artificial intelligence: an analytical review,'' \emph{Wiley
  Interdisciplinary Reviews: Data Mining and Knowledge Discovery}, vol.~11,
  no.~5, p. e1424, 2021.

\bibitem{longo2020explainable}
L.~Longo, R.~Goebel, F.~Lecue, P.~Kieseberg, and A.~Holzinger, ``Explainable
  artificial intelligence: Concepts, applications, research challenges and
  visions,'' in \emph{International Cross-Domain Conference for Machine
  Learning and Knowledge Extraction}.\hskip 1em plus 0.5em minus 0.4em\relax
  Springer, 2020, pp. 1--16.

\end{thebibliography}
\end{document}